\documentclass{article}

\PassOptionsToPackage{numbers, compress}{natbib}

\usepackage[final]{neurips_2025}
\usepackage{amsmath}

\usepackage[utf8]{inputenc} 
\usepackage[T1]{fontenc}    
\usepackage{hyperref}       
\usepackage{url}            
\usepackage{booktabs}       
\usepackage{amsfonts}       
\usepackage{nicefrac}       
\usepackage{microtype}      
\usepackage{xcolor}         
\usepackage{graphicx}
\usepackage{wrapfig}

\title{Meta-Learning Fourier Neural Operators for Hessian Inversion and Enhanced Variational Data Assimilation}

\author{
Hamidreza Moazzami\\
School of Computational Science and Engineering,\\McMaster University, Hamilton, Canada\\
\And
Asma Jamali\\
School of Computational Science and Engineering,\\McMaster University, Hamilton, Canada\\
\And
Nicholas Kevlahan \\
Mathematics and Statistics,\\McMaster University, Hamilton, Canada\\
\texttt{kevlahan@mcmaster.ca}
\And
Rodrigo A.~Vargas-Hernández \\
Department of Chemistry and Chemical Biology,\\McMaster University, Hamilton, Canada\\
\texttt{vargashr@mcmaster.ca}
}

\begin{document}

\maketitle

\begin{abstract}
    Data assimilation (DA) is crucial for enhancing solutions to partial differential equations (PDEs), such as those in numerical weather prediction, by optimizing initial conditions using observational data. Variational DA methods are widely used in oceanic and atmospheric forecasting, but become computationally expensive, especially when Hessian information is involved. To address this challenge, we propose a meta-learning framework that employs the Fourier Neural Operator (FNO) to approximate the inverse Hessian operator across a family of DA problems, thereby providing an effective initialization for the conjugate gradient (CG) method. Numerical experiments on a linear advection equation demonstrate that the resulting FNO-CG approach reduces the average relative error by $62\%$ and the number of iterations by $17\%$ compared to the standard CG. These improvements are most pronounced in ill-conditioned scenarios, highlighting the robustness and efficiency of FNO-CG for challenging DA problems.
\end{abstract}

\section{Introduction} \vspace{-0.3cm} 
Partial differential equation (PDE) models, such as those used in weather prediction, often rely on initial conditions that are unknown or only approximately known, which makes them a significant source of error \cite{magnusson2019dependence}. Data assimilation (DA) addresses this by optimally estimating the initial state using sparse, arbitrarily distributed observations in space and time. Widely applied in oceanic and atmospheric forecasting, DA methods are typically categorized as sequential or variational \cite{navon2009data, park2013data, de2022coupled}. The most common variational method, 4D-Var, frames DA as an optimal control problem, minimizing the discrepancy between model simulations and observations over time and space \cite{le1986variational}.

4D-Var optimization typically relies on gradient-based methods, but convergence and accuracy can be improved by incorporating Hessian information \cite{cartis2024convergent}. Direct computation of the Hessian, or even Hessian–vector products, is often prohibitively expensive, particularly in ill-conditioned settings. To mitigate this, several strategies have been explored, including preconditioning \cite{haben2011conditioning}, reduced-order modelling \cite{cstefuanescu2015pod}, and Hessian approximation methods \cite{cioaca2013efficient, solvik20254d}.

Here, FNOs are leveraged to improve the efficiency and accuracy of 4D-Var. FNOs, known for learning PDE operators efficiently \cite{li2020fourier, khetrapal2024numerical, choi2024applications}, have been applied to DA and PDE-constrained optimization problems. For instance, \citet{singh2024learning} combined FNO-based predictions with correction operators for DA, while \citet{kang2025adjoint} employed FNO surrogate solvers for PDEs to accelerate gradient computations. In this work, we reinterpret the FNO through the lens of meta-learning \cite{hospedales2021meta}, training it to model the inverse Hessian operator across a distribution of 4D-Var problems. In doing so, the FNO becomes a reusable, task-adaptive solver that shapes the optimization trajectory, accelerating convergence while retaining the rigour of classical solvers. This data-driven approach offers a computationally efficient alternative to traditional iterative schemes, where the model implicitly captures the structure of the inverse Hessian to advance the state of DA.

\section{Methodology} \vspace{-0.3cm}
\noindent
\textbf{4D-Var framework}. \; Variational DA methods are based on minimizing a cost function by tuning the model trajectory to the observations, which reduces cumulative errors over time. The time evolution of the model state from time step $k$ to $k+1$ is defined as, $\mathbf{u}_{k+1} = \mathcal{M}_{k+1}(\mathbf{u}_{k})$, where $\mathbf{u}$ is the model state-vector and $\mathcal{M}$ is a nonlinear model operator. The 4D-Var loss function is defined as, 
\begin{equation}\label{eq_2}
  J(\mathbf{u})= \frac{1}{2}(\mathbf{u}_0-\mathbf{u}^b)^T\mathbf{B}^{-1}(\mathbf{u}_0-\mathbf{u}^b) +
  \frac{1}{2}\sum_{k=0}^{K}(\mathbf{H}\mathbf{u}_k-\mathbf{u}^{obs}_k)^T\mathbf{R}^{-1}
   (\mathbf{H}\mathbf{u}_k-\mathbf{u}^{obs}_k)
\end{equation}
where the first term is a regularization term that minimizes the distance between the initial state, $\mathbf{u}_0$, and the background state, $\mathbf{u}^b$, which is the prior information about the study system.
$\mathbf{H}$ is the linear observation operator that interpolates the model forecasts into observations. $\mathbf{B}$ and $\mathbf{R}$ are the background error covariance matrix and the observation error covariance matrix, respectively.
The second term is simply the difference between observations, $\mathbf{u}_k^{obs}$ and the PDE simulation, $\mathbf{u}_k$, over time.
The gradient of the loss function is given by, 
\begin{equation}\label{eq_3}
\small
\nabla{J}(\mathbf{u}_0)= \mathbf{B}^{-1}(\mathbf{u}_0-\mathbf{u}^b)- \mathbf{H}_0^T\mathbf{R}_0^{-1}\mathbf{d}_0+\mathbf{M}_1^T\left [\mathbf{H}_1^T\mathbf{R}_1^{-1}\mathbf{d}_1+\mathbf{M}_2^T\left [\mathbf{H}_2^T\mathbf{R}_2^{-1}\mathbf{d}_2+...+\mathbf{H}_K^T\mathbf{R}_K^{-1}\mathbf{d}_K \right ] \right ],
\end{equation}  
where $\mathbf{d}_k$ is the distance between the observations and the model forecast, and is defined as,  
\begin{equation}\label{eq_4}
\mathbf{d}_k=(\mathbf{u}^{obs}_k-\mathbf{H}_k\mathcal{M}_{k}\mathcal{M}_{k-1}...\mathcal{M}_2\mathcal{M}_1\mathbf{u}_0), 
\end{equation}
where $\textbf{M}_k$ is the tangent linear model (TLM) of the nonlinear model, $\mathcal{M}_k$,  and $\textbf{M}_k^T$ is its adjoint model, which evolves the state backward in time \cite{Asch2016}. 
If the PDE is linear ($\textbf{M}_k$), the operators $\textbf{G}_k$ containing both $\textbf{H}$ and $\textbf{M}_k$ and $\textbf{G}_k^T$ containing $\textbf{H}^T$ and $\textbf{M}_k^T$ are defined as, 
\begin{equation}\label{eq_5}
    \mathbf{G}_k=\mathbf{H}_k\mathbf{M}_{k}\mathbf{M}_{k-1}...\mathbf{M}_2\mathbf{M}_1 \quad \text{and} \quad \mathbf{G}_k^T=\mathbf{M}_{1}^T\mathbf{M}_{2}^T...\mathbf{M}_{k-1}^T\mathbf{M}_k^T\mathbf{H}_k^T.
\end{equation} 

Since the model and the observation operator are linear and the cost function is quadratic, the Hessian of the loss function, $\nabla^2 J$ can be used to write the exact first-order Taylor expansion of the gradient, $\nabla {J}(\mathbf{u}_0)=\nabla {J}(\mathbf{u}_0=0)+\nabla^2 J\mathbf{u}_0$, where $\mathbf{u}_0$ is optimum at $\nabla {J}=0$. Therefore, the optimal initial state, $\mathbf{u}_0^{opt}$ is achieved via: 
\begin{equation}\label{eq_9}
    \nabla^2 J \; \mathbf{u}_0^{opt} = \mathbf{f}
\end{equation} 
where $\mathbf{f}$ is the solution of the gradient equation in Eq. (\ref{eq_3}) by setting $\mathbf{u}_0 = \mathbf{0}$,   
\begin{equation}\label{eq_10}
    \mathbf{f} = -\nabla{J}(\mathbf{u}_0=\mathbf{0}) = \mathbf{B}^{-1}\mathbf{u}^b+\sum_{k=0}^{K}\mathbf{G}^T_k\mathbf{R}_k^{-1}\mathbf{u}_k^{obs}.
\end{equation} 
Although Eq.~(\ref{eq_9}) is derived for a linear PDE, incremental 4D-Var extends it to nonlinear PDEs by updating the nonlinear model in the outer loop and using its TLM in the inner loop to define a convex quadratic cost minimized via the Gauss--Newton approximation \cite{neveu2011applications, lawless2005approximate}. This highlights the potential applicability of our approach to nonlinear PDEs.

\noindent
\textbf{FNO for Hessian-based Data Assimilation}.
The inverse of the Hessian in Eq. (\ref{eq_9}) is computationally intensive for high-dimensional systems and, therefore, is usually solved iteratively \cite{solvik20254d}. The CG method typically requires a large number of iterations to solve Eq. (\ref{eq_9}), especially when the condition number of the Hessian is high \cite{fisher2009data}. The multigrid method can be an efficient alternative approach for solving Eq. (\ref{eq_9}). However, it is restricted to elliptic problems, which limits its applicability \cite{debreu2016multigrid, neveu2011multigrid}. 

In this study, a surrogate FNO for the Hessian inverse operator is introduced, allowing Eq. (\ref{eq_9}) to be solved in a single step, without relying on an iterative process,
\begin{equation}\label{eq_11}
    \mathbf{u}_0^{opt} = [\nabla^2 J]_{\text{FNO}}^{-1} (\mathbf{f}).
\end{equation}  
In Eq. (\ref{eq_11}), the FNO model plays the role of the inverse Hessian, $[\nabla^2 J]_{\text{FNO}}^{-1}$, that maps $\mathbf{f}$ to the target variable, $\mathbf{u}_0$.  
$\mathbf{f}$ is computed by running the standard adjoint model on the observations, $\mathbf{u}_k^{\text{obs}}$ (Eq. (\ref{eq_10})). Recall that the observations are sparse and distributed arbitrarily in space and time. An important advantage of the proposed approach is that, unlike most previous ML-based DA studies \cite{he2020physics}, the sparse observations $\mathbf{u}_k^{\text{obs}}$ are implicit to the FNO, embedded in $\mathbf{f}$ and fed into the model, eliminating the need for a separate observation term in the loss and simplifying training. The FNO uses multiple spectral convolution layers that apply a learned transformation to low-frequency modes, efficiently capturing long-range spatial dependencies and maintaining flexibility across resolutions.

The FNO model was trained on 5,400 points using the Adam optimizer with a learning rate of $10^{-4}$ and a 32 batch size. The implementation was based on the \href{https://github.com/raj-brown/fourier_neural_operator}{\texttt{fourier\_neural\_operator}} library, with 16 Fourier modes in the architecture. All experiments were performed on an Apple M2 chip. 

\section{Experimental Setup} \vspace{-0.3cm}
The DA framework in this study approximates the initial condition $\textbf{u}_0$, of a linear advection equation solved on a periodic domain: 
\begin{equation}\label{eq_12}
  \frac{\partial u}{\partial t} + c \frac{\partial u}{\partial x} = 0, \quad 
  x \in \left[-\frac{x_{\max}}{2}, \frac{x_{\max}}{2}\right), \quad t \in [0, T]
\end{equation} 

where $x_{max}=100\:m,\: T=90\:s$, and $\: c=0.92\: m.s^{-1}\;$. The true initial state, $\mathbf{u}_0^T$, is obtained by perturbing the background state, $\mathbf{u}^b$, with spatially periodic noise, $\eta$, composed of a finite sum of modulated cosine functions with varying frequencies, $\mathbf{u}_0^T = \mathbf{u}^b + \mathbf{B}^{1/2} \; \eta$.
The FNO is trained with pairs of ($\mathbf{u}_0^T$, $\mathbf{f}$), which are generated by varying multiple factors. The general form of the background state is, 
\begin{equation}\label{eq_14}
    {u}^b(x) = 0.5+ \beta\sin(\alpha\frac{2\pi x}{x_{\max}}+\phi)
\end{equation} 
\noindent where \( \alpha \), \( \beta \), and \( \phi \) take the following values: $\alpha \in \{2, 4, 6\}$, $\beta \in \{0.1, 0.3, 0.5, 0.7, 1\}$, and $\phi \in \left\{ 0, \frac{\pi}{3}, \frac{\pi}{4} \right\}$.
The second-order autoregressive (SOAR) distribution is employed to model the background error covariance, $\mathbf{B}$ \cite{stewart2013data, ingleby2001statistical}:  
\begin{equation}\label{eq_15}
    B_{ij} = \sigma_b^2 \left(1 + \frac{D_{ij}}{L} \right) \exp\left( -\frac{D_{ij}}{L} \right),
\end{equation} 
where $D_{ij}$ is the distance between each pair of grid points, $\sigma_b^2$ is the variance of the background error and $L$ is the correlation length scale. To generate the training data, the correlation length scale $L$ is chosen from the set 
\( \{5\Delta x, 10\Delta x, 15\Delta x, 20\Delta x, 25\Delta x\} \), 
where \( \Delta x \) is the spatial grid's resolution. 

Based on the above assumptions for the parameters in \( \mathbf{u}^b \) and the correlation length scale \( L \) in \( \mathbf{B} \), 
different true initial state vectors \( \mathbf{u}_0^T \) were generated. 
Each \( \mathbf{u}_0^T \) can be paired with multiple \( \mathbf{f} \) vectors by considering various observation configurations, all sampled from the true solution. 
We assumed the number of spatial observations is selected from the set \( \{2, 4, 6, 8\} \),  with all observations equally spaced, and their temporal availability is specified by the set \( \{1, 4, 6, 10, 15, 20\} \),  where each value denotes the interval (in time steps) between consecutive observations. 
Samples were generated by discretizing the PDE in Eq. (\ref{eq_12}) with the Lax–-Wendroff scheme and computing $\mathbf{f}$ in Eq. \ref{eq_10} via its adjoint, with training on $[0, T]$ and testing on $[T, 2T]$, each containing $5,400$ samples. 

\begin{figure}[htp!]
    \centering
    \includegraphics[width=\linewidth]{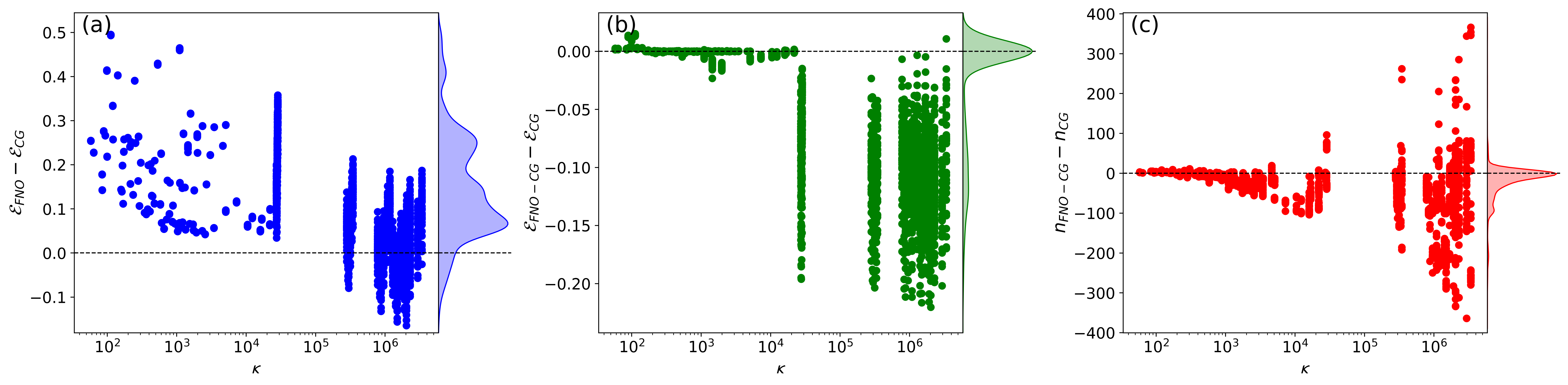}\vspace{-0.2cm}
    \caption{
    (a) Difference between the relative errors of the FNO and the CG model. (b) Difference between the relative errors of the FNO-CG and the CG model. (c) Difference in the number of iterations between the FNO-CG and the CG model. All panels are as a function of the condition number of the Hessian, $\kappa$.
    }
    \label{fig:rel_er_cond}
\end{figure}

\section{Results and Discussion}  \vspace{-0.3cm}
In this section, we demonstrate that although the $\mathbf{u}_0$ predicted with an FNO model, Eq.~(\ref{eq_11}), is typically less accurate than that of the standard CG solver, their combination improves results compared to using either method alone. To measure the difference between the true solution, $\textbf{u}_0^T$ and the predicted $\textbf{u}_0$, we used the relative error:
 \begin{equation}\label{eqn:relative_error}
     {\cal E}(\mathbf{u}_0) = \frac{||\mathbf{u}_0^T - \mathbf{u}_0||}{||\mathbf{u}_0^T||}.
 \end{equation}
Fig. \ref{fig:rel_er_cond}a presents the difference between the relative errors of the standalone FNO model and the CG method,  $\Delta {\cal E}_{\text{FNO}} = {\cal E}_{\text{FNO}} - {\cal E}_{\text{CG}}$, plotted against the condition number of the Hessian, $\kappa$, for the test dataset consisting of 5,400 samples. The distribution displayed to the right of Fig. \ref{fig:rel_er_cond}a illustrates that, for the majority of cases, $\Delta {\cal E}_{\text{FNO}}$ is positive, indicating that the standalone FNO model generally does not outperform the CG method. 

As previously discussed, the CG method is widely used in the DA framework, but its performance is highly sensitive to the choice of the initial guess: poor initialization can substantially slow convergence. To overcome this limitation, we propose using FNOs to provide an effective initialization for solving Eq.~(\ref{eq_9}), leading to the FNO-CG approach.
Table~\ref{tab:cg_initializers} compares the standard CG method, initialized with the background state $\mathbf{u}^b$ (Eq.~(\ref{eq_14})), against FNO-CG, where the initializer is the FNO-predicted state $\mathbf{u}_0$. The results show that FNO-CG consistently improves both accuracy and efficiency.
Fig.~\ref{fig:rel_er_cond}b presents the distribution of $\Delta {\cal E}_{\text{FNO-CG}} = {\cal E}_{\text{FNO-CG}} - {\cal E}_{\text{CG}}$. Most samples yield negative values, indicating smaller relative errors with FNO-CG. Cases with positive $\Delta {\cal E}_{\text{FNO-CG}}$ remain close to zero and are predominantly associated with well-conditioned problems, where CG alone is already sufficient. This underscores the robustness of FNO-CG in handling poorly conditioned DA problems.
Further, Fig.\ref{fig:rel_er_cond}c illustrates the difference in the number of iterations, $\Delta n_{\text{FNO-CG}} = n_{\text{FNO-CG}} - n_{\text{CG}}$, which is predominantly negative across the dataset, confirming the efficiency gain. On average, FNO-CG reduces the relative error by 62\% and the iteration count by 17\%.

\begin{table}[h!]
\centering
\caption{Comparison of CG, FNO, and FNO-CG results for the test data.}
\label{tab:cg_initializers} 
\begin{tabular}{lcc}
\hline
\textbf{} & \textbf{Average Relative Error} $\downarrow$ & \textbf{Average Total Iterations} $\downarrow$ \\
\hline
CG & $4.26\times 10^{-2}$ & $127$ \\
FNO-CG & $\mathbf{1.61\times 10^{-2}}$ & $\mathbf{105}$ \\
FNO & $1.86\times 10^{-1}$ & -- \\
\hline
\end{tabular}
\end{table}

\begin{figure}[ht!]
    \centering
    \includegraphics[width=\textwidth]{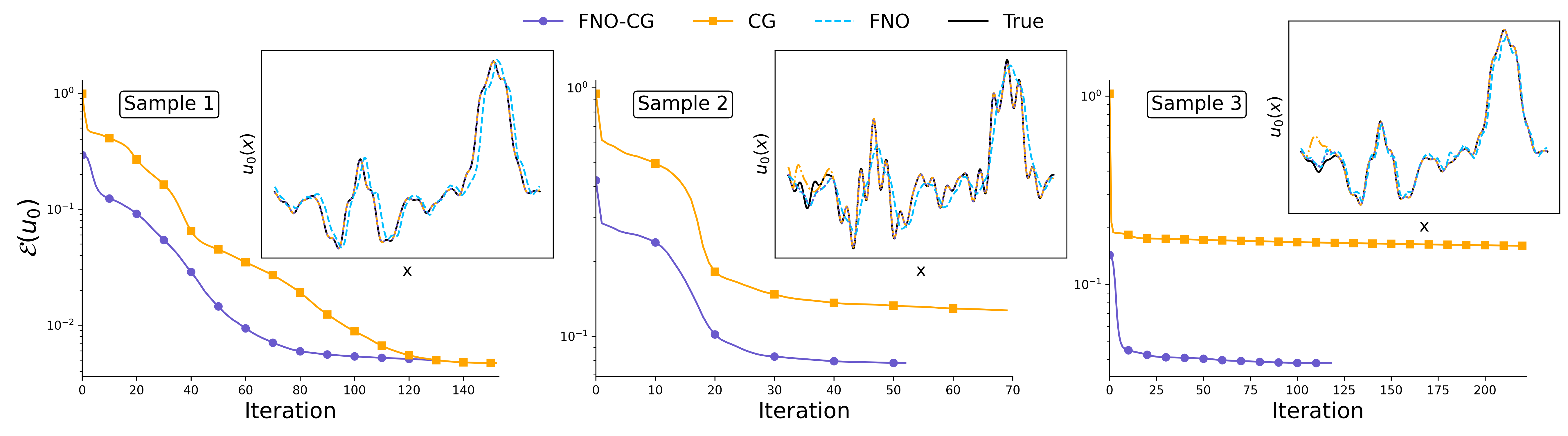}
    \caption{
    Main panels: relative error as a function of CG iterations when the initial $\mathbf{u}_0$ is set to the background state $\mathbf{u}_b$ or to the initialization predicted by the FNO model. Inset panels: predicted $\mathbf{u}_0$ obtained with CG, FNO, and FNO-CG methods for three randomly selected test samples.
    }
    \label{fig:relative_error}
\end{figure}

Fig.~\ref{fig:relative_error} compares CG, FNO, and FNO-CG on three randomly selected test samples. For Sample 1, CG and FNO-CG achieve similar accuracy, but FNO-CG converges with 24 fewer iterations. For Samples 2 and 3, FNO-CG attains $39\%$ and $76\%$ higher accuracy, while also reducing the iteration count by 17 and 103, respectively. The inset panels show the corresponding $\mathbf{u}_0$ predicted by each method. Although the standalone FNO model exhibits a relatively high average relative error, $1.86 \times 10^{-1}$ (see Table~\ref{tab:cg_initializers}), it still captures the overall structure of $\mathbf{u}_0$, providing a useful initialization for CG.

\newpage
\section{Conclusion} \vspace{-0.3cm}

We introduced a meta-learning framework based on Fourier Neural Operators to enhance variational data assimilation by approximating the inverse Hessian operator. The proposed FNO-CG method combines the predictive power of FNOs with the robustness of the classical CG algorithm, yielding substantial improvements: a $62\%$ reduction in average relative error and a $17\%$ decrease in iteration count compared to standard CG. Distribution analyses show that these gains are most pronounced in ill-conditioned problems, where optimization is typically more difficult.
Beyond this application, our results demonstrate that FNOs can approximate operators beyond PDE solvers, here targeting the inverse Hessian, a central object in many inverse problems. By learning the mapping from $\mathbf{f}$ to an effective initial state $\mathbf{u}_0$, FNO-CG reduces the need for expensive iterative procedures.
Future work will extend this meta-learning approach to nonlinear PDEs, with a particular focus on challenging chaotic systems such as the Kuramoto–Sivashinsky equation.

\section{Acknowledgment}
NKRK was supported by the Natural Sciences and Engineering Research Council of Canada (NSERC) under Grant no. RGPIN-2024-05282. 
RAVH acknowledges the support from the Digital Research Alliance of Canada and NSERC Discovery Grant No. RGPIN-2024-06594.

\bibliographystyle{unsrtnat}
\bibliography{bibliography}

\end{document}